\documentclass[letterpaper]{article} 
\usepackage{aaai24}  
\usepackage{times}  
\usepackage{helvet}  
\usepackage{courier}  
\usepackage[hyphens]{url}  
\usepackage{graphicx} 
\usepackage{natbib}  
\usepackage{caption} 
\usepackage{algorithm}
\usepackage{algorithmic}
\usepackage{xcolor}
\usepackage{amsmath}
\usepackage{color}
\usepackage{subcaption}

\newcommand{\KMFigure}[1]{Figure~\ref{#1}}
\newcommand{\KMEq}[1]{Eq.~(\eqref{#1})}

\definecolor{seagreen}{rgb}{0.18, 0.55, 0.34}
\definecolor{royalpurple}{rgb}{0.47,0.32,0.66}
\definecolor{brown(traditional)}{rgb}{0.59, 0.29, 0.0}
\definecolor{blue(traditional)}{rgb}{0.15, 0.29, 0.50}
\definecolor{azure}{rgb}{0, 0.5, 1}
\definecolor{kly}{rgb}{0, 0.18, 0.65}
\definecolor{lightred}{rgb}{0.83,0.28,0.28}

\usepackage{newfloat}
\usepackage{listings}
\DeclareCaptionStyle{ruled}{labelfont=normalfont,labelsep=colon,strut=off} 
\floatstyle{ruled}
\newfloat{listing}{tb}{lst}{}
\floatname{listing}{Listing}
\nocopyright

\title{FreePIH: Training-Free Painterly Image Harmonization with Diffusion Model}

\author{
    Ruibin LI\textsuperscript{\rm 1},
    Jingcai Guo\textsuperscript{\rm 1},
    Song Guo\textsuperscript{\rm 2},
    Qihua Zhou\textsuperscript{\rm 1},
    Jie Zhang\textsuperscript{\rm 1}
}
\affiliations{
    \textsuperscript{\rm 1} Department of Computing, The Hong Kong Polytechnic University\\
    \textsuperscript{\rm 2} Department of Computer Science and Engineering, Hong Kong University of Science and Technology\\
    ruibin.li@connect.polyu.hk, jingcai.guo@gmail.com, songguo@cse.ust.hk, csqzhou@comp.polyu.edu.hk, jie-comp.zhang@polyu.edu.hk
}

\usepackage{bibentry}

\begin{document}

\maketitle

\begin{abstract}

This paper provides an efficient training-free painterly image harmonization (PIH) method, dubbed \textit{FreePIH}, that leverages only a pre-trained diffusion model to achieve state-of-the-art harmonization results. 
Unlike existing methods that require either training auxiliary networks or fine-tuning a large pre-trained backbone, or both, to harmonize a foreground object with a painterly-style background image, our \textit{FreePIH} tames the denoising process as a plug-in module for foreground image style transfer. 
Specifically, we find that the very last few steps of the denoising (i.e., generation) process strongly correspond to the stylistic information of images, and based on this, we propose to augment the latent features of both the foreground and background images with Gaussians for a direct denoising-based harmonization. 
To guarantee the fidelity of the harmonized image, we make use of multi-scale features to enforce the consistency of the content and stability of the foreground objects in the latent space, and meanwhile, aligning both fore-/back-grounds with the same style. 
%
Moreover, to accommodate the generation with more structural and textural details, we further integrate text prompts to attend to the latent features, hence improving the generation quality. 
Quantitative and qualitative evaluations on COCO and LAION 5B datasets demonstrate that our method can surpass representative baselines by large margins.
\end{abstract}

\section{Introduction}


Image compositing is a fundamental task in image editing that enables users to combine foreground and background images to create new artwork. Oftentimes, the foreground and background images have varying colors and structures. Our objective is to seamlessly integrate a foreground item into a different background image, producing a natural and visually pleasing result. Over the years, there has been a growing interest in image compositing, with researchers exploring various methods to enhance the quality of composite images\cite{VGG-DP1,VGG-DP2,AAAIcomposite,VGG-DP4}. The overall framework utilized in these studies remains largely unchanged. Typically, a set of loss terms (such as content loss, style loss, and stability loss) is designed based on a pre-trained feature extractor network, often VGG-19\cite{vggnet}. These loss terms are then iteratively optimized to update the pixel values of the foreground image. One notable advantage of this framework is its avoidance of the need for additional data collection or resource-intensive model training or fine-tuning processes. Consequently, it can serve as a plug-in module built upon an off-the-shelf pre-trained model.

Recent advancements in image generation techniques have led researchers to explore the use of Text to Image Diffusion Models (T2I-DM) for image compositing\cite{LDM}. T2I-DM, combined with CLIP\cite{CLIP}, enables users to generate images based on natural language prompts. However, one of the challenges faced with T2I-DM is the loss of control over the generated images. In scenarios where users want to composite specific items into an image and describe them using natural language, T2I-DM may not generate the desired output or may be difficult to guide using simple words. Previous work, such as Dreambooth\cite{Dreambooth}, can inject specific items into output images, but this approach requires lengthy fine-tuning processes. Other works like text-driven editing approaches are insufficient for image compositing as sometimes it is hard to give accurate verbal representations to capture the details or preserve the identity and appearance of a given object image\cite{BlendDM}.

\begin{figure}[t]
\centering
\includegraphics[width=0.45\textwidth]{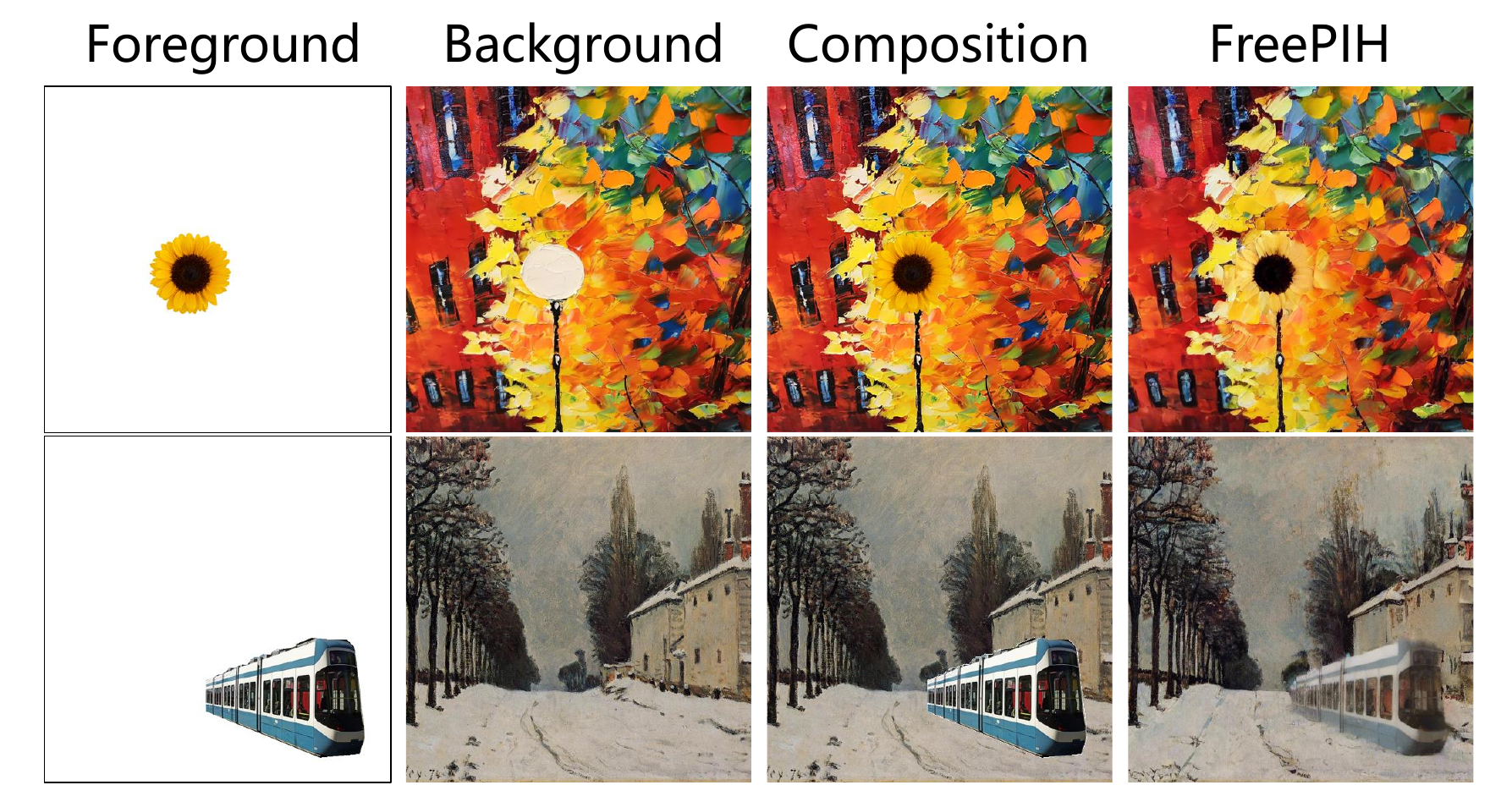} 
\caption{Example of painterly image harmonization with our proposed FreePIH.}
\label{Figure:example}
\end{figure}




In order to enable T2I-DM with image compositing capabilities, a straightforward solution involves using T2I-DM to generate the background image and then applying an image compositing algorithm to blend the foreground item into the generated image. However, this approach often results in a foreground image that fails to capture the style of the background, leading to unnatural fused results. Additionally, previous approaches typically employ VGG-19 as the feature extractor, necessitating the use of both the neural network module employed by the T2I-DM and VGG-19 for compositing. However, we find that the modules within T2I-DM already serve as effective multi-scale feature extractors. Unlike VGG-19, which is trained for classification, the modules in T2I-DM are specifically trained for image generation and possess the ability to extract multi-scale feature maps, thereby surpassing VGG-19 in handling the complex feature representations required for image editing.

In this paper, we take advantage of the pre-trained T2I-DM to perform image compositing and introduce a method named \textit{FreePIH}. The background can either be the images generated by the T2I-DM or provided by the user. This approach differs from text-driven editing with diffusion models, as it allows users to add items to specific locations while preserving their structure and appearance, providing greater control over their AIGC (Artificial Intelligence Generated Content) artworks. Specifically, our framework takes a tuple ($\mathbf{x}_G$, $\mathbf{x}_I$, $m$) that represents the background, foreground, and mask, respectively. We utilize the Variational Autoencoder (VAE) image encoder in T2I-DM to extract low-level content feature maps. 

To obtain high-level style feature maps, we first apply data augmentation based on the diffusion forward process to the latent features extracted by the VAE. The augmented latent features then interact with the text input in the DM module using a cross-attention mechanism. The resulting feature map serves as our high-level style representation. To ensure the consistency between input and output images, the level of forward noise injection in the data augmentation needs to be controlled. Through iterative optimization, we seamlessly blend the foreground into the background. Importantly, our method eliminates the need for model fine-tuning, distinguishing it from existing baselines, particularly DM-based approaches. Furthermore, our method empowers T2I-DM users with increased control over their AIGC artworks.


Our contributions can be summarized as follows:

\begin{itemize}
    \item We propose FreePIH, which can work as a plug-in module to enable image harmonization on off-the-shelf T2I-DM without the need for collecting new data, training auxiliary networks, and fine-tuning pre-trained models.
    \item We conduct noise augmentation on the latent features and leverage the corresponding output to accurately capture stylistic information based on the feature of DM.
    \item With the composition capacity of FreePIH, users gain enhanced autonomy in shaping their AIGC artworks when using DM-based genetative model.
    \item Qualitative and quantitative analysis reveals that our \textit{FreePIH} method can generate more natural fusion images compared with other baselines.
\end{itemize}

\section{Related work}

\subsection{Image Compositing}

Image compositing has long been a challenging task in the field of image editing, with the aim of seamlessly integrating a given foreground image into a target background image. Most of the current work is based on the framework designed by \cite{DP-framework}. This framework utilizes a pretrain neural network model to extract multi-scale feature maps, which are then used to calculate a set of loss terms such as style loss, content loss, and stability loss. Subsequent works have focused on modifying the design of these loss terms and the feature extractor. 

Some works, such as DPH\cite{DPH} and DIB\cite{DIB}, optimize the optimization process by incorporating different loss terms like poisson image loss and histogram loss. Others, like PHDNet\cite{AAAIcomposite} and FDIT\cite{FreqDomain}, have found that transforming the feature maps into the frequency domain can improve appearance preservation and compositing harmonization. In addition, CNN-based networks such as RainNet \cite{VGG-DP2}, DoveNet \cite{CNN1}, and BAIN \cite{VGG-DP1}, as well as attention-based networks like SAM \cite{Attention1} and CDTNet \cite{VGG-DP4}, have been employed to replace the former VGG-19 feature extractor. Different from the above work, we exploit the untapped potential of the off-the-shelf T2I-DM models to leverage their image harmonization capabilities. Since the modules within the T2I-DM models are trained on a large-scale dataset for image generation, the features extracted by these modules possess innate competence as zero-shot multi-scale feature extractors for image compositing and harmonization tasks.

\subsection{Text to Image Diffusion}



The Diffusion Model (DM)\cite{DDPM} is an innovative AI model that draws inspiration from non-equilibrium thermodynamics. It operates by defining a Markov chain of diffusion steps, gradually introducing random noise to data, and then learning to reverse the diffusion process in order to generate desired data samples from the noise. In the case of image generation, DM generates a random Gaussian noise image and progressively removes noise in a step-by-step manner until a clear image is obtained. 

By integrating with the CLIP text encoder, DM gains the ability to utilize natural language prompts to guide the diffusion generation process. Models like Stable Diffusion\cite{LDM}, DallE-\cite{Dalle2}, and Midjourney have demonstrated the remarkable ability to perform text-to-image (T2I) guided generation. However, while T2I-DM serves as a powerful tool for image generation, there are instances where the generated images may not align with the user's expectations. For example, if the desired prompt is ``\textit{dog sitting in front of a door}'', providing T2I-DM with the text prompt ``\textit{dog}'' might yield an entirely different image with a dog sitting elsewhere. Hence, providing users with more control over their AI-generated artworks remains a challenge.

\begin{figure*}[htbp]
\centering
\includegraphics[width=0.87\textwidth]{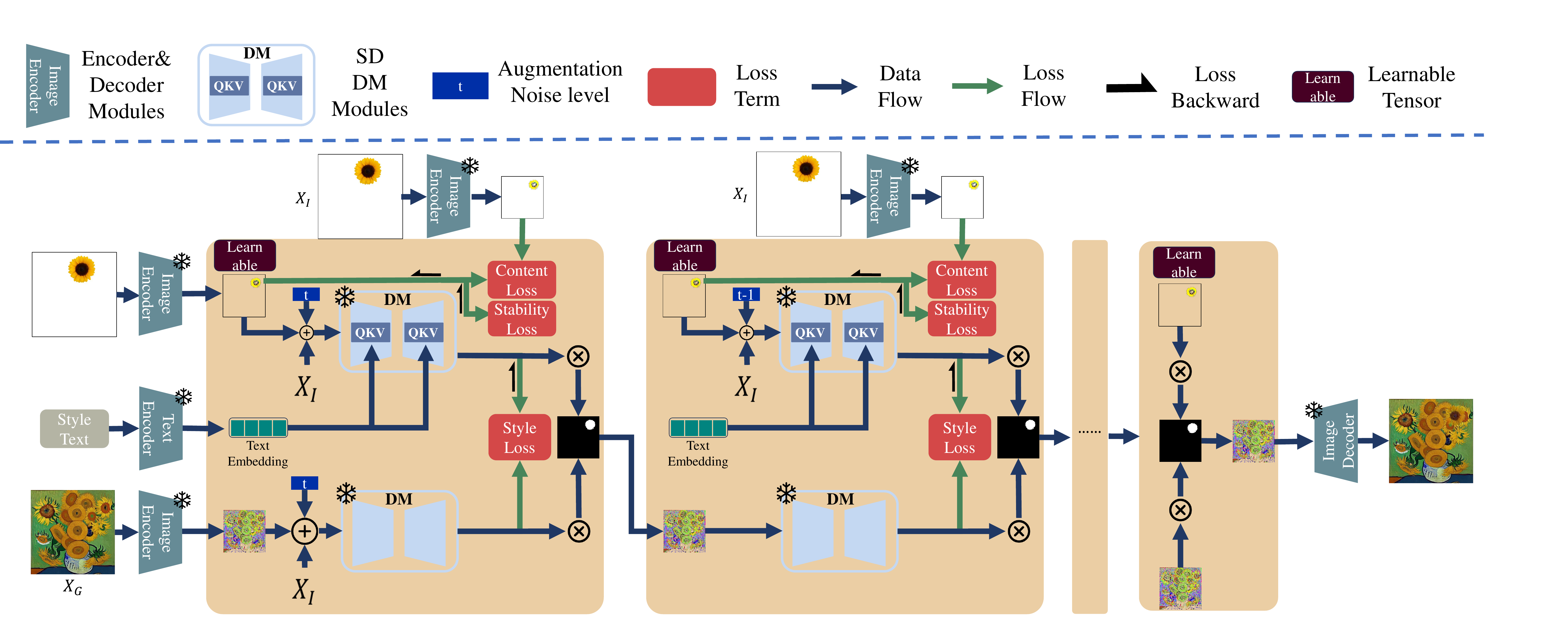} 
\caption{The architecture of our FreePIH, we modify the pre-train stable diffusion model and add several loss terms to control the style transformation of foreground items.}
\label{Figure:architecture}
\end{figure*}

\subsection{Guide Diffusion}
To enhance the control of DM, researchers have proposed several updated versions of T2I-DM. Works such as Dreambooth\cite{Dreambooth} and text-inversion\cite{textinversion} address this issue by tailoring T2I-DM generation to users' personalized requirements. However, these methods necessitate extensive time and computational resources, as they involve hours of fine-tuning a pre-trained T2I-DM model. Additionally, Dreambooth requires a significant amount of images with similar semantic content, which may not always be readily available.

An alternative approach, known as prompt-to-prompt, facilitates image editing by modifying the cross-attention modules of T2I-DM\cite{prompt2prompt}. However, the edited images generated through this method are limited to those produced by T2I-DM itself, as the attention editing operation relies on the previous attention feature map. Consequently, it is not ideal for editing user-provided images.

Recent research endeavors have attempted to equip T2I-DM with text-driven editing capabilities\cite{BlendDM}. Nevertheless, accurately and concisely describing a personalized demand can be challenging in some cases. Furthermore, even when an appropriate text prompt is provided (e.g., a precise description of a dog's ears, eyes, and nose features), T2I-DM may capture these features but ultimately generate a dog with completely different characteristics. This diverges from the original intention of having the dog appear in the desired background.


\section{Method}



Given an input image, denoted as $\mathbf{x}_G$, which can be either provided by the user or generated by the T2I-DM model using a semantic prompt $d$, our objective is to blend this image with another user-provided image, $\mathbf{x}_I$, using a binary mask $m$. The goal is to create a fused image, $\mathbf{x}_F$, where the content in the masked region, $\mathbf{x}_F \circ m$, closely resembles the structure and texture of the original image $\mathbf{x}_I$. In other words, we want $\mathbf{x}_F \circ m$ to be approximate to $\mathbf{x}_I$ (denoted as element-wise multiplication, $\circ$). Additionally, the unmasked region should remain unchanged, meaning $\mathbf{x}_F \circ (1-m) = \mathbf{x}_G$. It is also important to ensure that the blended region $\mathbf{x}_F \circ m$ and unchangeable region $\mathbf{x}_F \circ (1-m)$ have a consistent style, resulting in a seamless and natural transition between the two regions.

To achieve this, we propose a method that utilizes the intermediate results from T2I-DM to guide the style transfer of the user-provided image. 
This is done by introducing a blending loss that consists of content loss, style loss, and stability loss. 

\subsection{Overall Workflow}


As depicted in \KMFigure{Figure:architecture}, we leverage a pre-train T2I-DM architecture for our task. We have found that the VAE and Unet module in T2I-DM serve as excellent feature extractors. Prior to compositing the foreground and background images, we utilize the VAE image encoder to convert $\mathbf{x}_{G}$ and $\mathbf{x}_{I}$ into latent features $\hat{\mathbf{x}}_{G}$ and $\hat{\mathbf{x}}_{I}$. Additionally, we initialize a learnable latent feature $\hat{\mathbf{x}}_{L}$ with the same value as $\hat{\mathbf{x}}_{I}$ as a starting point. The objective of our method is to optimize the learnable feature $\hat{\mathbf{x}}_{L}$ so that it seamlessly integrates with $\hat{\mathbf{x}}_{G}$ while preserving the majority of the features observed in $\hat{x}_{I}$. To achieve this goal, we introduce noise into the latent features $\hat{\mathbf{x}}_{G}$ and $\hat{\mathbf{x}}_{L}$. These augmented latent features are then fed through the DM network, and the resulting output is used to calculate the style loss. By performing backpropagation with both content loss and stability loss, we are able to update the learnable feature $\hat{\mathbf{x}}_{L}$. We denote the updated result as $\hat{\mathbf{x}}^{t-1}_{L}$. Next, we update $\hat{\mathbf{x}}_{G}$ using the input mask, which can be expressed as follows:

\begin{equation}
    \hat{\mathbf{x}}_{G}^{t-1} = DM_{\theta}(\hat{\mathbf{x}}_{G}^t) \circ (1-m) + DM_{\theta}(\hat{\mathbf{x}}_{L}^{t-1}) \circ m,
    \label{Equation:xg}
\end{equation}
where the $\hat{\mathbf{x}}^{t-1}_{L}, \hat{\mathbf{x}}_{G}^{t-1}$ then work as the input for next iteration. We repeat this process until $t=0$. Finally, we have $\hat{\mathbf{x}}_{F}=\hat{\mathbf{x}}_{G}^{0}$, and we can use the VAE image decoder to decode the latent feature $\hat{\mathbf{x}}_{F}$ to $x_{F}$, which is our final output. Note that in the compositing process, only the learnable parts are updated, while the DM, and VAE modules are frozen. Additionally, the compositing process can be completed within a few seconds, whereas fine-tuning a T2I-DM with Dreambooth may require as long as a day. This streamlined approach allows for efficient and timely image compositing while minimizing the overall computational burden.

\subsection{Augmentation and Denoising}

\begin{figure}[htbp]
\centering
\includegraphics[width=\linewidth]{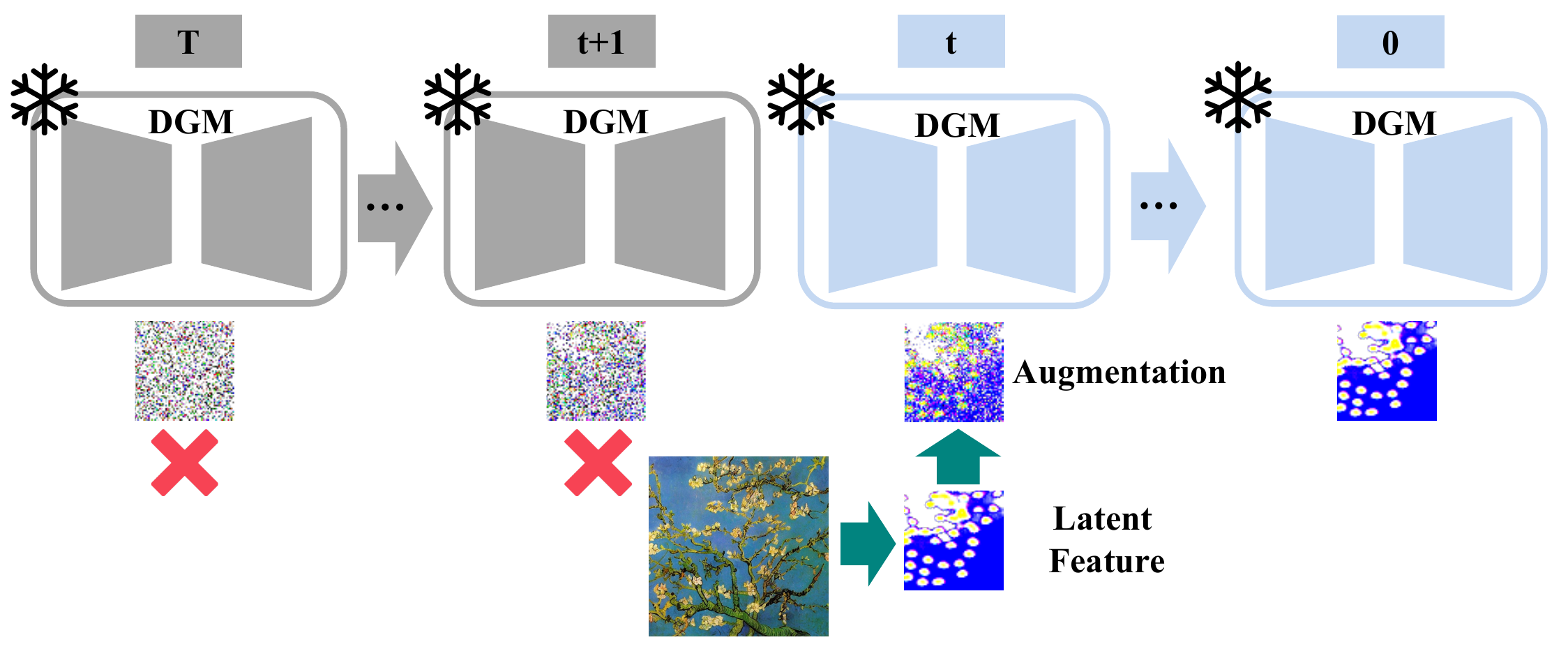} 
\caption{Conduct the noise augmentation only on the last steps of the denoising process. We avoid the denoising calculation from $T$ to $t+1$.}
\label{Figure:augmentation}
\end{figure}


To capture the stylistic information with T2I-DM, we first conduct the augmentation over the input latent feature $\hat{x}_L,\hat{x}_G$ by injecting noise into the input latent feature. The results distribution after noise augmentation is as follows:
\begin{equation}
    q(\hat{\mathbf{x}}_L^t \vert \hat{\mathbf{x}}_L) = \mathcal{N}(\hat{\mathbf{x}}_L^t; \sqrt{\bar{\alpha}_t} \hat{\mathbf{x}}_L, (1 - \bar{\alpha}_t)\mathbf{I}),
\end{equation}
where $t$ is a hyperparameter that controls the noise inject level. Typically, we set $t$ to be less than 0.2 times the total denoising steps ($T$), as depicted in Figure \ref{Figure:augmentation}. This decision is based on three key observations. Firstly, previous studies \cite{prompt-to-prompt,DM-partation} have demonstrated that the denoising process in DM can be divided into two parts. After generating an image from a purely noisy image $\mathbf{x}^T$, the majority of the content remains unchanged from the first part. Subsequently, in the remaining denoising steps, the details and style are gradually refined. Therefore, to capture the style features, we only need to introduce a small amount of noise to modify the style. Secondly, our observations indicate that injecting excessive noise can disrupt the structure of the original image. Since the denoising process involves inherent randomness, injecting too much noise can lead to a loss of control over the image content and result in a different background. Thirdly, by ensuring that $t$ is less than $T$, we can expedite the inference process. If we inject $t$ steps of noise, we only need to denoise the same number of steps, thereby accelerating the overall process.


The denoising steps follow the normal DM process which we have:
\begin{equation}
\begin{aligned}
    &q(\mathbf{x}^{t-1}|\mathbf{x}^{t},c) = \mathcal{N}(\mathbf{x}^{t-1}| \boldsymbol{\mu}_{\theta},\sigma^2 \mathbf{I}), \\
    & \boldsymbol{\mu}_{\theta} = \frac{1}{\sqrt{\alpha_t}} \Big( \mathbf{x}^t - \frac{1 - \alpha_t}{\sqrt{1 - \bar{\alpha}_t}} \boldsymbol{\epsilon}_\theta(\mathbf{x}^t, t, c) \Big), \\
    & \sigma^2 = \frac{1 - \bar{\alpha}_{t-1}}{1 - \bar{\alpha}_t} \cdot \beta_t,
\end{aligned}
\label{Equation:denoise}
\end{equation}
where $\boldsymbol{\epsilon}_\theta$ represents a neural network that takes the noised image $\mathbf{x}^t$, the time index $t$, and additional conditions as inputs, and predicts the noise that should be removed from $\mathbf{x}^t$. 
The solver used to sample $\mathbf{x}^{t-1}$ in \KMEq{Equation:denoise} can be any solver proposed by previous works, such as DDPM\cite{DDPM}, DDIM\cite{DDIM}, DPM++\cite{DPMsolver}, and so on.

Since the output of DM is already the noise version of the input (i.e., $\mathbf{x}^{t-1}=DM(\mathbf{x}^t,t)$), we can directly update $\hat{\mathbf{x}}_G$ using \KMEq{Equation:xg}, which is shown as the intermediate line in \KMFigure{Figure:architecture}. The updated value can then serve as the input for the next iteration. As for $\hat{\mathbf{x}}_L$, we retain the noise injection steps with different $t$.



\subsection{Loss}


The fundamental concept behind style transfer is to utilize multiple loss functions to achieve a balance between various objectives, including transferring the visual style from image $\mathbf{x}_I$ to the style of image $\mathbf{x}_G$, preserving the structure and details of the input image $\mathbf{x}_I$, and seamlessly merging $\mathbf{x}_I$ into $\mathbf{x}_G$. The overall loss function can be represented as follows:
\begin{equation}
    \mathcal{L} = \omega_{sty} \mathcal{L}_{sty} + \omega_c \mathcal{L}_c + \omega_{sta} \mathcal{L}_{sta}.
\end{equation}
The loss functions for the mentioned objectives, denoted as $\mathcal{L}_{sty}, \mathcal{L}_c, \mathcal{L}_{sta}$, are used to measure and optimize the model's performance. These objectives represent different aspects and are assigned weights represented by $\omega_{sty}, \omega_c, \omega_{sta}$ to ensure a balanced combination of the losses. The distinguishing factor among various works lies in their formulation of these loss functions and the feature extractors employed. In our research, we take a different approach from previous studies that utilize VGG-Net as the feature extractor. Instead, we capitalize on the modules present in the T2I-DM, which themselves serve as exceptional multi-scale feature extractors. This allows us to harness the features extracted by these modules for the calculation of different loss functions, thus enhancing the overall performance of the model.

\subsubsection{Style Loss}
To obtain the style feature of the background, denoted as $\hat{\mathbf{x}}_G$, we subject it to augmentation and input it into the DM. The output serves as the style feature representation of the background. To better utilize the multi-modal features of T2I-DM, we go a step further for the foreground. We incorporate textual information to provide DM with knowledge about the foreground item. This is achieved by encoding the text $c$ using the CLIP text encoder. The encoded text then guides the denoising process through the cross-attention mechanism.
\begin{equation}
    Attn = Softmax \Big( \frac{(\mathbf{W}_Q E_{img})(\mathbf{W}_K E_{txt})^T}{\sqrt{d}} \Big) \mathbf{W}_V E_{txt},
    \label{Equation:attention}
\end{equation}
where $E_{img}$ is the intermediate feature of DM and $E_{txt}$ is the embedding result of CLIP text encoder. $\mathbf{W}_Q,\mathbf{W}_K,\mathbf{W}_V$ are attention weights. Finally, we calculate the style loss as:
\begin{equation}
    \mathcal{L}_{sty} = || G(DM_{\theta}( \hat{\mathbf{x}}_L^t,t,c )) - G(DM_{\theta}( \hat{\mathbf{x}}_G^t,t,c ))  ||^2,
\end{equation}
where $G(\cdot)=DM_{\theta}(\cdot) DM_{\theta}(\cdot)^T \in \mathbf{R}^{N \times N}$ is the Gram matrix. The advantage of using the Gram matrix is that it can remove the location impact on the style representation. Meanwhile, the product of DM feature and its transposition can turn the local statistics feature into a global feature\cite{DPH}.

\subsubsection{Content Loss.} Since the latent feature $\hat{\mathbf{x}}_L$ serves as a comprehensive multi-scale content feature representation, we can calculate the content loss by measuring the difference between $\hat{\mathbf{x}}_L$ and $\hat{\mathbf{x}}_I$ as follows:

\begin{equation}
    \mathcal{L}_c = || \hat{\mathbf{x}}_L - \hat{\mathbf{x}}_I ||.
\end{equation}
By minimizing this loss term, we can ensure the content in the corresponding position of the target output have the close structure and detail as to the user providing $\mathbf{x}_I$.

\subsubsection{Stability Loss}
To increase the stability of the output and reduce the ambiguity during the generation process, we add histogram loss\cite{Hist} and total variation loss\cite{TV} into $\mathcal{L}_{sta}$. Histogram loss is calculated by:
\begin{equation}
    \mathcal{L}_{his} = || \hat{\mathbf{x}}_L - \mathbf{R}(\hat{\mathbf{x}}_L) ||^2,
\end{equation}
where $\mathbf{R}(\hat{\mathbf{x}}_L)=histmatch(\hat{\mathbf{x}}_L,\hat{\mathbf{x}}_G)$ is the histogram-remapped feature map by match $\hat{\mathbf{x}}_L$ to $\hat{\mathbf{x}}_G$.

Total variation loss is calculated by:
\begin{equation}
    \mathcal{L}_{tv} = \sum_{i,j} (\hat{\mathbf{x}}_L(i,j) - \hat{\mathbf{x}}_L(i,j-1))^2 + (\hat{\mathbf{x}}_L(i,j) - \hat{\mathbf{x}}_L(i-1,j))^2,
\end{equation}
where $\hat{\mathbf{x}}_L(i,j)$ represent the feature in the position $(i,j)$.

Finally, we have:
\begin{equation}
    \mathcal{L}_{sta} = \lambda_{his} \mathcal{L}_{his} + \lambda_{tv} \mathcal{L}_{tv},
\end{equation}
where $\lambda_{his},\lambda_{tv}$ are two hyperparameters to balance the influence of these two loss terms. $\mathcal{L}_{sta}$ term can improve the compositing result by producing smoother output.

\subsection{Optimization}

For the optimization process, we have carefully selected the values of the weighting parameters: $\omega_{sty}$ is set to $1e7$, $\omega_{c}$ is set to $1e1$, and $\omega_{sta}$ is set to $1$. Additionally, we perform 5 rounds of optimization in each iteration, and instead of using the Adam solver, we utilize a quasi-Newton solver called L-BFGS to minimize the loss function instead of Adam solver as we found failed to composite the foreground image into the background with the same number of optimization rounds as the L-BFGS solver. 



\subsection{Second Stage Refinement.}

In order to enhance the quality of the fusion image and minimize artifacts in the transition areas between the foreground and background images, we employ a square mask that encompasses the entirety of the foreground region along with a portion of the background region. Subsequently, we introduce a slight amount of noise into this region and employ the same T2I-DM to denoise it. During this stage, we eliminate all loss terms and solely retain the text prompt, making the denoising process akin to the SDEdit process.


\begin{algorithm}[tb]
\caption{FreePIH}
\label{alg:algorithm}
\textbf{Input}: Foreground $\mathbf{x}_I$, Background $\mathbf{x}_G$, Mask $m$, Style text $c$, Pre-train $T2I-DM_{\theta}$, Augmentation strength $t$\\
\textbf{Output}: Fused image $\mathbf{x}_F$
\begin{algorithmic}[1] 
\STATE Let ENC be the VAE encoder of $T2I-DM_{\theta}$.
\STATE Let DEC be the VAE decoder of $T2I-DM_{\theta}$.
\STATE Let $DM$ be the Unet of $T2I-DM_{\theta}$.
\STATE $\hat{\mathbf{x}}_G = ENC(\mathbf{x}_G),\hat{\mathbf{x}}_I = ENC(\mathbf{x}_I)$.
\STATE $\hat{\mathbf{x}}_L = \hat{\mathbf{x}}_I$.
\STATE $Optimizer = LBFGS(\hat{\mathbf{x}}_L)$
\FOR{$i \in [t,...,0]$}
\STATE $\hat{\mathbf{x}}_L^{t} = Augmentation(\hat{\mathbf{x}}_L,t)$
\STATE $\hat{\mathbf{x}}_G^{t} = Augmentation(\hat{\mathbf{x}}_G,t)$
\STATE $\mathcal{L}_{sty} = StyleLoss(\hat{\mathbf{x}}_L^{i-1},\hat{\mathbf{x}}_G^{i-1})$
\STATE $\mathcal{L}_{c} = MSELoss(\hat{\mathbf{x}}_L,\hat{\mathbf{x}}_G)$
\STATE $\mathcal{L}_{sta} = StabilityLoss(\hat{\mathbf{x}}_L)$
\STATE $\mathcal{L} = \omega_{sty} \mathcal{L}_{sty} + \omega_c \mathcal{L}_c + \omega_{sta} \mathcal{L}_{sta}$
\STATE $Backward(\mathcal{L})$
\STATE $Step(Optimizer)$
\STATE $\hat{\mathbf{x}}_G = DM( \hat{\mathbf{x}}_G^i,i,c ) \circ (1-m) + DM( \hat{\mathbf{x}}_L^i,i,c ) \circ m$
\ENDFOR
\STATE $\hat{\mathbf{x}}_F = \hat{\mathbf{x}}_G$
\STATE $\mathbf{x}_F = DEC(\hat{\mathbf{x}}_F)$
\STATE \textbf{return} $\mathbf{x}_F$.
\end{algorithmic}
\end{algorithm}

\section{Experiment}

The foreground item and mask are generated from COCO, a large-scale object detection, segmentation, and captioning dataset. The background images are randomly selected from LAION 5B, a large-scale dataset with high visual quality images and captions scraped from the web.

\subsection{Baselines}

For comparison, we choose six baseline methods including non-DM-based Poisson image editing (PIE)\cite{PIE}, Deep Image Blending (DIB)\cite{DIB}, PHDNet\cite{AAAIcomposite} and DM-based SDEdit\cite{SDEdit}, SD-Text\cite{LDM}, and BlendDM\cite{BlendDM}. Among these, SD-Text needs detailed textual information about output images while other DM-based methods only need simple textual input about the foreground items. All of the codes and pre-train weights are released by the authors. The detailed descriptions of all these baselines are shown in the supplementary materials.

\begin{figure*}[thbp]
\centering
\includegraphics[width=\textwidth]{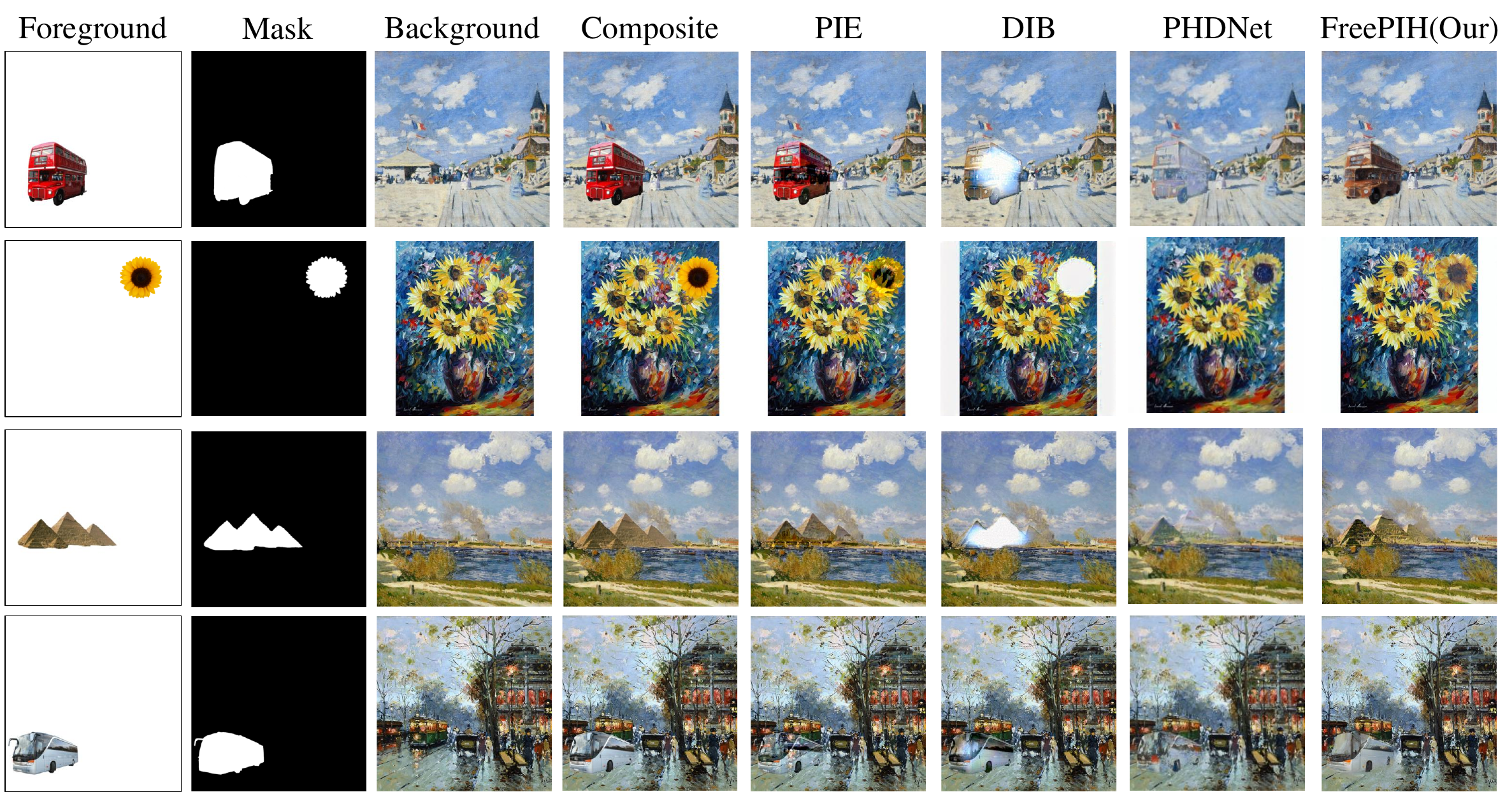} 
\caption{Example results of non-DM-based painterly image harmonization baselines and our FreePIH.}
\label{Figure:NDFqualitative}
\end{figure*}


For our FreePIH, we use the pretrain model sd-v1-5 released by Stability AI, all the parameters of every SD modules are frozen during our inference and we only update the latent feature of foreground content so we are actually training-free. All the methods are tested on a desktop equipped with a 12900K Intel CPU @3.20GHz, 64GB memory, and an NVIDIA RTX 4090 GPU.

\subsection{Quantitative Evaluation}

\begin{figure}[htp]
  \centering
  
  \begin{subfigure}{0.49\linewidth}
    \includegraphics[width=\textwidth]{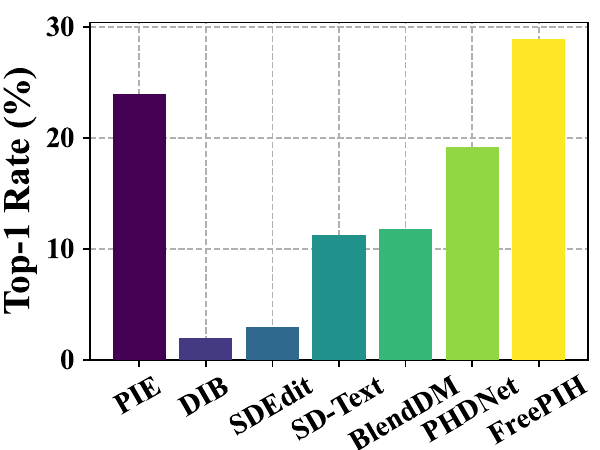}
  \end{subfigure}
  \begin{subfigure}{0.49\linewidth}
    \includegraphics[width=\textwidth]{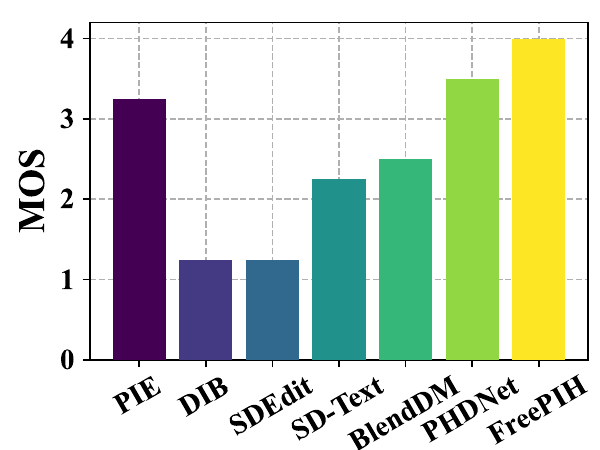}
  \end{subfigure}%
  \caption{(Left) The average Top-1 rate of all the baselines. (Right) The mean opinion score of different baselines, with a score from one (Bad) to five (Excellent). }
  \label{Figure:user_study}
\end{figure}

As mentioned in previous research, accurately calculating common metrics such as MSE, PSNR, and SSIM for harmonized images is challenging, particularly when foreground and background images are randomly selected. To overcome this limitation and evaluate performance quantitatively, we conducted a user study. Initially, we randomly selected 100 foreground and background pairs from the COCO\cite{COCO} and LAION 5B datasets\cite{LAION}, respectively. We then employed our method, along with all the baselines, to generate the composition images. Utilizing a subset of these images, we created a questionnaire in which each question consisted of the same images but produced by different methods. Users were asked to vote for the top-1 harmonious image, and a portion of the participants also rated the images on a scale from one (poor) to five (excellent). In total, we collected 365 votes from 73 users. Afterward, we summarized the average top-1 rate and calculated the Mean Opinion Score (MOS) based on the users' scoring\cite{MOS}. The results are depicted in Figure \KMFigure{Figure:user_study}, which clearly demonstrate that our FreePIH method outperforms the other baselines in our human evaluation test.

\subsection{Qualitative Evaluation}

The qualitative evaluation of non-diffusion-based baselines is presented in \KMFigure{Figure:NDFqualitative}. In our experiment, PIE maintains the original foreground feature after optimization; however, the optimized foreground occasionally becomes transparent. Consequently, the foreground content fails to obscure the background image, resulting in the visibility of background people, which appears unnatural. In some instances, DIB is unable to complete the transformation, as evident in the first three rows. Moreover, certain foreground regions are heavily influenced by the background style, leading to a clear separation between different areas. For example, in row 4, the center of the bus exhibits discordant blue and appears brighter than the rest. While PHDNet performs better compared to the previous baselines, sometimes the foreground becomes too harmonious, causing the loss of its original texture and details. This is apparent in the first and third rows, where the bus and pyramid assume the color of the sky. Particularly in row 3, the other two pyramids blend into the sky, leaving only one pyramid noticeable at first glance. In contrast, our FreePIH effectively maintains the texture and structure of both the foreground and background, seamlessly fusing them. For instance, FreePIH converts the original deep red to a red similar to the house in row 1, and successfully optimizes the boundary while preserving the colorful foreground items in rows 2-4. Unlike PIE, DIB, and PHDNet, our FreePIH does not require specific foreground images. Methods such as PIE, DIB, and PHDNet depend on extracting foreground items from separate images. Consequently, if the foreground image, like the sunflower, is a distinct item with no additional background, these methods fail to style-transform the foreground items.

\begin{figure}[thbp]
\centering
\includegraphics[width=\linewidth]{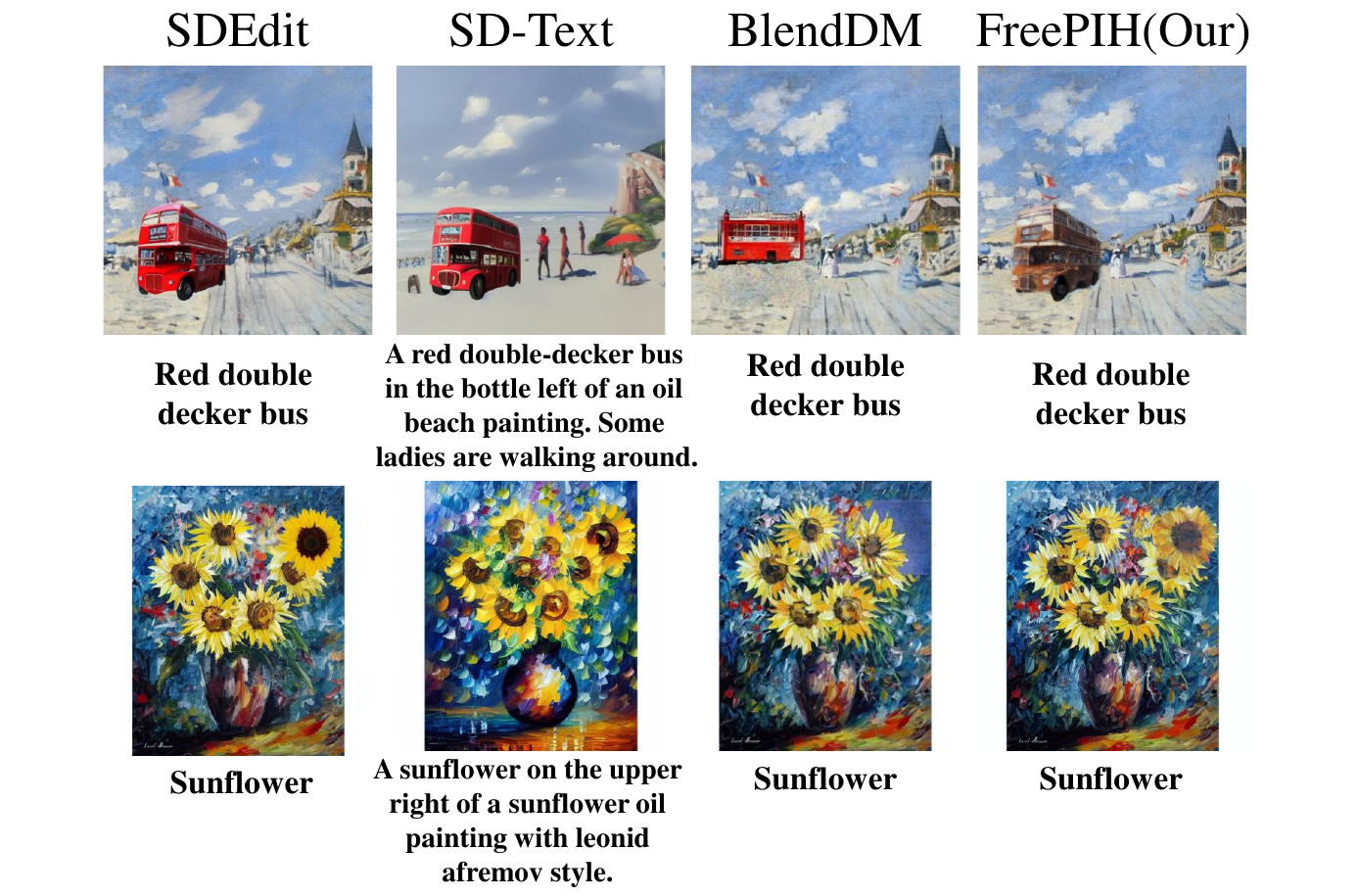} 
\caption{Example results of DM-based painterly image harmonization baselines and our FreePIH.}
\label{Figure:Qualitative_DM}
\end{figure}


The qualitative evaluation of diffusion-based baselines is depicted in \KMFigure{Figure:Qualitative_DM}. In each baseline, a text prompt is inputted to enhance the fidelity of the generated image. For SDEdit, BlendDM, and our FreePIH, we simply input the names of the foreground items. However, for SD-Text, our aim is to obtain output images that closely resemble our original image, therefore we need to provide more detailed text prompts, such as ``\textit{A sunflower on the upper right of a sunflower oil painting with Leonid Afremov style}'', in order to achieve a similar image. 

As illustrated in the figure, SDEdit successfully maintains the texture of the foreground items, but fails to transform their style. SD-Text is able to produce images with coherent semantic content as guided by the text prompt, though the foreground and background may deviate significantly from our expectations even when providing detailed textual descriptions of the original image. In contrast, BlendDM has the capability to introduce a new item into the designated mask region, but the item generated may not align with the one we originally inputted. On the other hand, our FreePIH generates harmonious fusion images with well-preserved foreground and background elements.

\subsection{Ablation Study}

\begin{figure}[thbp]
\centering
\includegraphics[width=\linewidth]{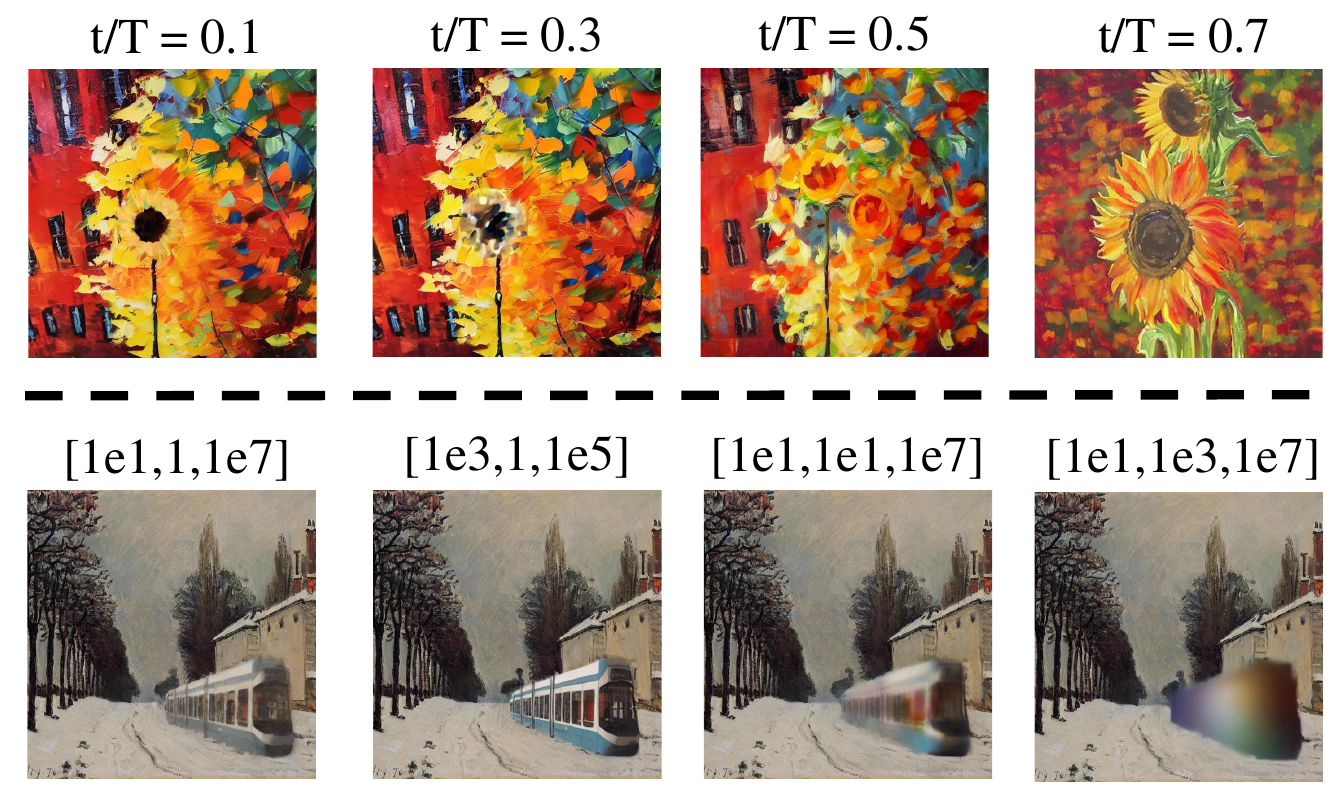} 
\caption{Ablation results of noise step, loss weight.}
\label{Figure:ablation}
\end{figure}

To evaluate the impact of certain hyperparameters, we conducted ablation experiments on the noise level and loss weight. The results of these experiments are shown in \KMFigure{Figure:ablation}. In the first row of the figure, we adjusted the noise level (represented by t/T) for the same harmonization task. We added different levels of loss in the forward process. As depicted in the figure, when the noise level is small, we were able to successfully fuse the foreground sunflower with the background painting. However, as the noise level increased, the texture and details of the foreground sunflower gradually deteriorated. Ultimately, when t/T reached values less than or equal to 0.5, the foreground became unrecognizable sunflowers. In the second row, we evaluated the impact of loss weight. Through empirical analysis, we found that the best weights to achieve the desired outcome were as follows: $\omega_{c}=1e1,\omega_{sta}=1,\omega_{sty}=1e7$. Increasing the weight of the content loss and decreasing the weight of the style loss resulted in better preservation of the original texture and details, but led to less harmonious output images. It is important to note that the stability loss should not be set too high, as it may cause the different parts of the foreground to become overly consistent. 

\begin{figure}[thbp]
\centering
\includegraphics[width=\linewidth]{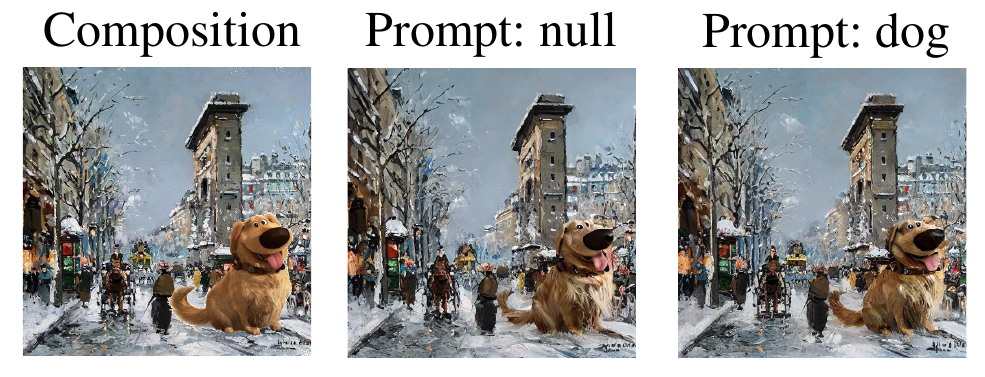} 
\caption{Text prompt can improve the fidelity of the foreground content.}
\label{Figure:ablation_text}
\end{figure}

Since the pre-trained Deep Matching (DM) models are optimized based on text conditions, providing an accurate text input can also help improve the fidelity of the foreground. In \KMFigure{Figure:ablation_text}, we illustrated the difference between an empty string text input and a prompt like ``\textit{dog}''. When the input text is empty, there may be distortions in certain parts of the foreground, such as the dog's eyes. However, inputting the prompt ``\textit{dog}'' helps eliminate these distortions.


\section{Conclusion}


In this paper, we present a pioneering approach to painterly image harmonization using diffusion-based techniques without the need for training. Our method leverages the observation that the final stages of the diffusion generation process capture crucial stylistic information in images. By utilizing the output features of the Diffusion Model (DM), we achieve a seamless transformation of foreground styles into background styles, resulting in harmonious image compositions. Notably, our method stands out from other baselines, particularly DM-based approaches, as it eliminates the requirement for extensive fine-tuning or training auxiliary modules on new data. It can be conveniently employed as a plug-in module for existing stable diffusion frameworks. Through both quantitative and qualitative evaluations, we demonstrate the superiority of our proposed FreePIH.

\bibliography{aaai24}

\end{document}